\documentclass[conference]{IEEEtran}

\usepackage{adjustbox}
\usepackage[ruled, vlined]{algorithm2e}
\usepackage{amsfonts}
\usepackage{amsmath}
\usepackage{amssymb}
\usepackage{amsthm}
\usepackage{array}  %
\usepackage{authblk}
\usepackage{booktabs}
\usepackage{capt-of}
\usepackage[font={small}]{caption}
\usepackage{subcaption}
\usepackage{colortbl}
\usepackage{inconsolata}  %
\usepackage{graphicx}
\usepackage{listings}
\usepackage{multicol}
\usepackage{multirow}
\usepackage[numbers]{natbib}
\usepackage{paralist}
\usepackage{pifont}
\usepackage{times}
\usepackage{tabularx}
\usepackage{thmtools}  %
\usepackage{wrapfig}
\usepackage{xcolor}
\usepackage{svg}

\definecolor{flodarkpurple}{rgb}{0.288,0.1196,0.7}

\definecolor{ganeshamber}{rgb}{1.0, 0.75, 0.0}

\newcommand{\etal}{\emph{et al.}}

\newcommand{\coolname}{\textit{ConceptFusion}}

\newcommand{\fg}{\textbf{f}^{\hspace{0.1em} G}} %
\newcommand{\fl}{\textbf{f}^{\hspace{0.1em} L}} %
\newcommand{\fs}{\textbf{f}^{\hspace{0.1em} P}} %

\usepackage[pagebackref=true,breaklinks=true,colorlinks,citecolor=flodarkpurple,bookmarks=true]{hyperref}

\begin{document}

\def\anonymized{0}  %

\title{\coolname{}: Open-set Multimodal 3D Mapping}

\if\anonymized1
    \author{Author Names Omitted for Anonymous Review. Paper-ID: \textbf{98}}
    \newcommand{\webpage}{https://sites.google.com/view/conceptfusion}
\else
    \definecolor{flodarkpurple}{rgb}{0.288,0.1196,0.7}
    \newcommand{\authorhref}[3][flodarkpurple]{\href{#2}{\color{#1}{#3}}}%

    \newcommand{\webpage}{https://concept-fusion.github.io/}

    \makeatletter
    \renewcommand\AB@affilsepx{, \protect\Affilfont}
    \makeatother

    \makeatletter
    \def\@maketitle{%
      \vskip 2em%
      \begin{center}%
      \let \footnote \thanks
        {\Huge \@title \par}%
        \vskip 1.5em%
        {\large
          \lineskip .5em%
          \begin{tabular}[t]{c}%
            \baselineskip=12pt
            \@author
          \end{tabular}\par}%
        \vskip 1em%
      \end{center}%
      \par
      \vskip 1.5em}
    \makeatother

    \author[1]{\authorhref{https://krrish94.github.io}{Krishna Murthy Jatavallabhula}}
    \author[2,$\dagger$]{\authorhref{https://www.alihkw.com/}{Alihusein Kuwajerwala}}
    \author[3,$\dagger$]{\authorhref{https://georgegu1997.github.io/}{Qiao Gu}}
    \author[4,$\dagger$]{\authorhref{https://scholar.google.com/citations?user=jFH3ShsAAAAJ&hl=en}{Mohd Omama}}
    \author[1]{\authorhref{https://taochenshh.github.io/}{Tao Chen}}
    \author[1]{\authorhref{https://www.csail.mit.edu/person/alaa-maalouf}{Alaa Maalouf}}
    \author[1]{\authorhref{https://people.csail.mit.edu/lishuang/}{Shuang Li}}
    \author[7,$\ddagger$]{\authorhref{https://epiception.github.io/}{Ganesh Iyer}}
    \author[8]{\authorhref{https://saryazdi.github.io/}{Soroush Saryazdi}}
    \author[5]{\authorhref{https://nik-v9.github.io/}{Nikhil Keetha}}
    \author[1]{\authorhref{https://ayushtewari.com/}{Ayush Tewari}}
    \author[1]{\authorhref{http://web.mit.edu/cocosci/josh.html}{Joshua B. Tenenbaum}}
    \author[6]{\authorhref{https://celsodemelo.net/}{Celso Miguel de Melo}}
    \author[4]{\authorhref{https://robotics.iiit.ac.in/}{K. Madhava Krishna}}
    \author[2]{\authorhref{http://liampaull.ca}{Liam Paull}}
    \author[3]{\authorhref{http://www.cs.toronto.edu/\~florian/}{Florian Shkurti}}
    \author[1]{\authorhref{https://groups.csail.mit.edu/vision/torralbalab/}{Antonio Torralba}}
    
    \affil[1]{\href{https://www.mit.edu}{MIT}}
    \affil[2]{\href{https://montrealrobotics.ca}{Universit\'e de Montr\'eal}}
    \affil[3]{\href{https://robotics.utoronto.ca/}{University of Toronto}}
    \affil[4]{\href{https://robotics.iiit.ac.in/}{IIIT Hyderabad}}
    \affil[5]{\href{https://www.cmu.edu}{CMU}}
    \affil[6]{\href{https://www.arl.army.mil/}{DEVCOM Army Research Lab}}
    \affil[7]{\href{https://www.aboutamazon.com/what-we-do/devices-services}{Amazon}}
    \affil[8]{\href{}{Concordia University}}

    \affil[$\dagger$]{Co-second authors}
    \affil[$\ddagger$]{Work done prior to Amazon}
\fi

\makeatletter
\let\@oldmaketitle\@maketitle
\renewcommand{\@maketitle}{\@oldmaketitle
\centering
\includegraphics[width=.95\linewidth]{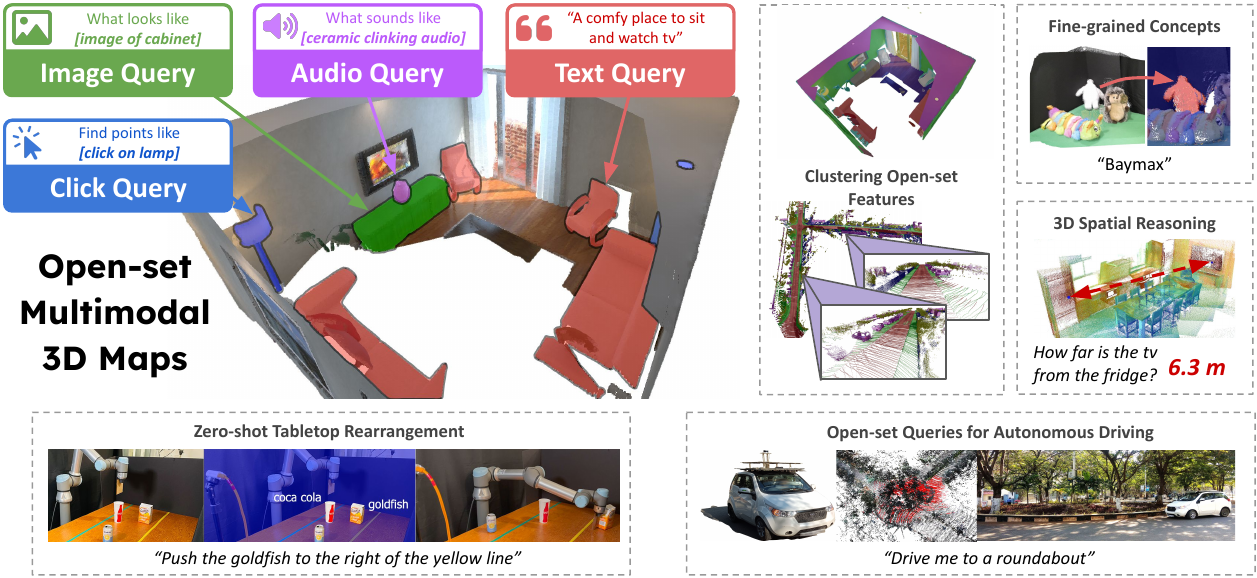}
\captionof{figure}{\textbf{\coolname{}} presents an approach to build \emph{open-set multimodal 3D maps} from RGB images and depth estimates from additional sources such as depth/stereo cameras or lidar, and features from foundation models like CLIP, DINO, AudioCLIP, etc.
These maps are built online, and can be queried for arbitrary concepts specified as text, images, audio samples, or clicks on the 3D map.
The fused features have an implicit understanding of semantic concepts, as evident by visualizing clusters obtained from a K-means algorithm.
\coolname{} features are significantly more adept at retaining fine-grained concepts, such as the disney character ``\emph{Baymax}''.
We also build 3D spatial reasoning modules that enable reasoning about frequently observed spatial relationships.
We demonstrate the applicability of \coolname{} to the real-world robotic tasks of tabletop manipulation of novel objects, and an urban autonomous driving setting. (\href{\webpage}{Webpage})
}
\label{fig:splash}
}

\makeatother

\maketitle

\begin{abstract}

Building 3D maps of the environment is central to robot navigation, planning, and interaction with objects in a scene. Most existing approaches that integrate semantic concepts with 3D maps largely remain confined to the closed-set setting: they can only reason about a finite set of concepts, pre-defined at training time.
Further, these maps can only be queried using class labels, or in more recent work, using text prompts.

We address \textit{both} these issues with \coolname{}, a scene representation that is: (i) fundamentally open-set, enabling reasoning beyond a closed set of concepts (ii) inherently multi-modal, enabling a diverse range of possible queries to the 3D map, from language, to images, to audio, to 3D geometry, all working in concert. \coolname{} leverages the open-set capabilities of today's foundation models that have been pre-trained on internet-scale data to reason about concepts across modalities such as natural language, images, and audio. We demonstrate that pixel-aligned open-set features can be fused into 3D maps via traditional SLAM and multi-view fusion approaches. This enables effective zero-shot spatial reasoning, not needing any additional training or finetuning, and retains long-tailed concepts better than supervised approaches, outperforming them by more than 40\% margin on 3D IoU. We extensively evaluate \coolname{} on a number of real-world datasets, simulated home environments, a real-world tabletop manipulation task, and an autonomous driving platform. We showcase new avenues for blending foundation models with 3D open-set multimodal mapping. We encourage the reader to view the demos on our project page: \href{\webpage}{\texttt{\webpage}}

\end{abstract}

\IEEEpeerreviewmaketitle

\setcounter{figure}{1} %

\section{Introduction}

One of the major catalysts of continued progress in 3D spatial perception~\cite{slam++,semanticfusion,d3vo,3dsg} has been the advent of deep convolutional networks trained on large datasets of images. Most of these advancements have relied on a \emph{closed-set} of concepts, a fixed set of labels available at training time. More recently, however, deep learning is witnessing yet another wave of advancements, this time with the emergence of increasingly larger and multimodal models trained on internet-scale data comprising billions of images, text, and audio~\cite{fmodels}.
Foundation models like CLIP~\cite{clip}, DINO~\cite{dino}, AudioCLIP~\cite{audioclip}, and their variants have shown impressive performance on \emph{open-set} scenarios, where the concepts of interest are supplied only at inference time. In this work, we bridge the gap between the rich open-set capabilities enabled by large foundation models and the semantic reasoning abilities expected of futuristic 3D mapping systems.

To be as broadly applicable as possible to a diverse set of robotics tasks, map representations need to be usable zero-shot (i.e. without the need to be retrained each time reasoning capabilities for a new task are desired), and must also posess the following two capabilities:
first, \textbf{3D maps should be open-set}; they should capture a large variety of concepts (orders of magnitude more than existing systems), and at varying levels of detail. For example, the concept ``\emph{can of soda}'' could equivalently be ``\emph{something to drink}'' or a ``\texttt{<particular brand of soda>}'' or ``\emph{a refreshment}''.
Second, \textbf{3D maps should be multimodal}; they should be queryable using as many modalities as robots or end-users can leverage. 
For instance, the search for a particular object in a map should work equally well if the query is a single \underline{word} (e.g. ``\emph{soda}''), a longer \underline{sentence} with additional context (e.g. ``\emph{is there a can of soda on the kitchen table?}), an \underline{image} of a soda can, or just the `pop' \underline{sound} associated with opening a soda can.

Foundation models possess some of the desired traits needed to achieve open-set, multimodal representations, but are not directly applicable to 3D mapping. This major limitation exists because most foundation models consume images (e.g., CLIP~\cite{clip}, ALIGN~\cite{align}, AudioCLIP~\cite{audioclip}) and produce only a single vector encoding of the entire image in an embedding space.
On the other hand, recent approaches trained specifically to align foundation features to 2D pixels \emph{forget} a large number of concepts during finetuning~\cite{dont-stop-learning-clip} (see Fig.~\ref{fig:pixel-aligned-qualitative}).
This does not allow for the level of precise (pixel-level or object-level) reasoning robotic perception systems need across a wide range of concepts, particularly for interaction with the external 3D world (e.g., navigation, manipulation).

To this end, we propose \coolname{}; an open-set and multimodal 3D mapping technique that blends advances in foundation models for images, language, and audio, with advances in dense 3D reconstruction and simultaneous localization and mapping (SLAM). We demonstrate that pixel-level foundation features may be fused into 3D maps by leveraging precisely the same surface fusion techniques as for fusing depth or color information into a 3D map. Crucially, we show that this approach is conceptually simple, principled, and effective even in the zero-shot setting (requiring no additional training or finetuning of foundation model features). In addition, these features can be queried using computationally efficient vector similarity metrics.
Our key contributions are the following:
\begin{compactitem}
\item An approach to open-set multimodal 3D mapping that constructs map representations queryable by text, image, audio, and click queries in a zero-shot manner.
\item A novel mechanism to compute pixel-aligned (local) features from foundation models that can only generate image-level (global) feature vectors. This is a key prerequisite for 3D mapping, and our approach captures long-tailed concepts significantly better than supervised or finetuned counterparts, outperforming them by a large margin ($>$ 40\% mIoU).
\item A new RGB-D dataset, UnCoCo, to evaluate open-set multimodal 3D mapping. UnCoCo comprises 78 common household/office objects tagged with more than 500K queries across modalities.
\end{compactitem}
We evaluate \coolname{} on multiple real-world datasets and tasks, including searching for objects in the real world and simulated home environments, robot manipulation tasks, and autonomous driving.

\section{Related work}

\begin{figure*}[!thb]
\centering
\includegraphics[width=\textwidth]{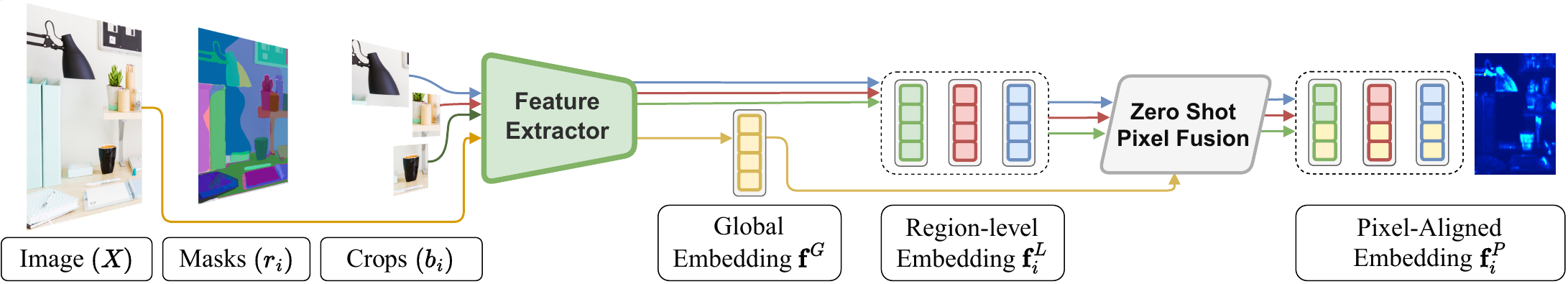}
\caption{\coolname{} constructs \textbf{pixel-aligned features}
$\fs$ by: processing input images to generate generic (class-agnostic) object masks (regions) $r_i$,
computing a bounding box for each region and extracting a local feature vector $\fl_i$,
computing a global feature $\fg$ for the input image as a whole,
and fusing the region-specific features with global features as illustrated in Fig.~\ref{fig:fusionmath} and described in Sec.~\ref{sec:fusing_features}.
}
\vspace{-0.45cm}
\label{fig:pipeline}
\end{figure*}

\textbf{Foundation models}~\citep{fmodels} are trained on vast quantities of data in a self-supervised fashion and accomplish a wide array of tasks, without the need for additional re-training or finetuning.
Image-language aligned models like CLIP~\cite{clip} and ALIGN~\cite{align} encode text and images into a shared concept space, and have driven progress in several open-set tasks~\cite{dalle,dalle2,styleclip,vqganclip,pointe,gu2021open,ding2022decoupling, openseg}.
They have also been extended to other modalities, such as audio~\cite{audioclip,wav2clip} and video~\cite{clip-vip}.
In a similar vein, features from image-only foundation models like DINO \cite{dino} have been employed as drop-in solutions for reasoning about visual concepts, such as classification, detection, segmentation, keypoint estimation~\cite{dinofts, dinofts2}.

Models like CLIP only align concepts to text or images as a whole; and not to image pixels or regions~\cite{regionclip}. This has led to a series of approaches that extract \textbf{pixel-aligned features}~\cite{lseg, openvsp, openseg, ovseg, regionclip}\footnote{We use the term \emph{pixel-aligned} to refer to features that can be attributed to every pixel in the input image (unlike CLIP~\cite{clip} features, which are global image descriptors). These features are typically computed at a coarser resolution, and upsampled via nearest neighbor interpolation, to match the original image resolution.} that address poor localization quality by training or fine-tuning a pixel-aligned model over a labeled dataset.
LSeg~\cite{lseg} leverages pixel-level labels from segmentation datasets, while OpenSeg~\cite{openseg}, OVSeg~\cite{ovseg}, and RegionCLIP~\cite{regionclip} use image-caption datasets and employ region-word grounding.
However, as we show in Sec.~\ref{sec:compute_features}, while this improves performance over concepts present in the finetuning dataset, these models do poorly at recalling concepts infrequent in the label set used for finetuning.
To retain knowledge of all concepts captured by the underlying CLIP model, MaskCLIP~\cite{maskclip} proposes a zero-shot approach that combines self-attention with engineered $1 \times 1$ convolution layers for open-set segmentation.
However, it struggles with delineating object boundaries, and long-tailed concepts, as we show in Sec~\ref{sec:compute_features} and Sec~\ref{sec:uncoco-eval}.
Our proposed pixel-aligned features mitigate all of the aforementioned issues by operating over region proposals and by aligning features computed across regions,
enabling a variety of spatial reasoning applications.

To fuse our pixel-aligned features into a 3D map, we leverage approaches pioneered by the \textbf{dense 3D mapping} community.
Approaches to dense SLAM estimate camera motion, scene geometry, and optionally, color from RGB(-D) images.
At the core of nearly all of these methods is the volumetric fusion technique from Curless and Levoy~\cite{curless1996volumetric}, which has been adapted for real-time incremental capture of surface geometry and color~\cite{kinectfusion, pointfusionbase, kintinuous, elasticfusion, dai2017bundlefusion}.
In this work, we extend this line of work to additionally fuse pixel-aligned foundation features to 3D maps in a conceptually simple and computationally efficient manner.
Noteworthy research efforts in bridging the gap between \emph{online} 3D mapping and open-set concept acquisition include ~\cite{semantic-paintbrush, ilabel}. In contrast, our work leverages large foundation features learned \emph{offline}, over billions of data samples.

Recent approaches exploring \textbf{foundation models for robotics} have demonstrated impressive open-set interaction abilities~\cite{robotlangsurvey,expglang,saycan2022arxiv,rt12022arxiv,li2022pre}.
While most of these approaches focus on planning and control, we provide a complementary perspective; that of perception systems that build explicit models of scenes and are able to query for concepts.
In Sec.~\ref{sec:outlook}, we present our outlook for how both of these classes of approaches can augment each other.

\section{Concurrent work}

Concurrently, multiple approaches are beginning to interface 2D foundation features for 3D scene understanding.

CLIP-Fields encodes a 3D map and pixel (or region)-aligned foundation features (LSeg~\cite{lseg}, Detic~\cite{detic}, Sentence-BERT~\cite{sentence-bert}) into a compact neural network~\cite{clipfields}.
This scene-specific neural network acts as a queryable database which aligns image and language embeddings with 3D scene points, and can be applied to open-set queries specified in language.
A new \textit{CLIP-Field} is trained per scene; and it remains to be explored whether learned CLIP-Fields could potentially generalize to new scenes, or to changes within a scene.

VLMaps~\cite{huang23vlmaps}, LM-Nav~\cite{shah2022robotic}, CoWs~\cite{gadre2022clip}, and NLMap-Saycan~\cite{chen2022nlmapsaycan} leverage the open-set features from pixel-aligned LSeg~\cite{lseg} models for robot navigation based on language commands.
More recently, OpenScene~\cite{openscene} applied pixel-aligned LSeg~\cite{lseg} and OpenSeg~\cite{openseg} to open-set 3D segmentation; demonstrating that features from these 2D extractors can be distilled into neural networks operating over 3D data.
Ding~\etal{}~\cite{ding2022language} additionally distill features from more complex 2D open-set tasks, such as image captioning, into a model that consumes 3D data.
A fundamental point-of-difference with the above set of approaches is that, we demonstrate that it is possible to construct 2D and 3D features zero-shot, alongside the mapping process, and without requiring any finetuning or distillation.
This zero-shot capability is also a key enabler in terms of our superior performance on long-tailed concepts and complex queries, and alignment with other modalities.

Another recent approach by Mazur~\etal{}~\cite{FeatureRealisticFusion} performs real-time fusion of features from foundation models for images, such as DINO~\cite{dino}, into compact 3D neural maps. Our approach, while open-set in this spirit, extends to multiple modalities such as language and audio; presenting a novel method to extract per-pixel features from CLIP.

Perhaps the closest approach to ours is semantic abstraction by Ha and Song~\cite{semantic-abstraction}, which also proposes a zero-shot approach to computing 3D-aligned CLIP features.
They explore a different mechanism (attention-explainability~\cite{attention-explainability}) to extract relevant regions corresponding to a text query; and demonstrate a complementary set of capabilities (completion of partly-observed objects, localizing hidden objects from language descriptions). We refer the interested reader to~\cite{semantic-abstraction} for more details.
In this work, we focus on unconditionally assigning CLIP features to image pixels and subsequently to 3D maps; over a wide variety of 3D perception and robotics tasks.
A recently proposed approach, 3D-CLR~\cite{3dclr}, also proposes neural operators for concept learning over 3D scenes. This inspires our design of 3D spatial query modules, however with a significant difference: we lift foundation model features to 3D in a zeroshot manner, preserving the full range of capabilities the models are equipped with; while ~\cite{3dclr} relies on LSeg~\cite{lseg}.

\section{The \coolname{} approach}

\textbf{The open-set multimodal 3D mapping problem}: Given a sequence of image (and depth) observations of an environment $\mathcal{I}=\{I_t\}$ ($t \in \{0 \cdots T\}$), we build an open-set multimodal 3D map $\mathcal{M}$.
This map is \emph{queryable} for concepts from multiple modalities, using query vectors $q_{\textnormal{mode}} \in \mathbb{R}^d$. Multidimensional signals such as images, text, audio, and clicks can be encoded into such a vector space using a modality-specific encoder (a foundation model) $\mathcal{F}_{\textnormal{mode}}$.

We first present a general feature fusion technique, extending traditional dense mapping approaches to incorporate per-pixel features in addition to color and depth information.
We then present our algorithm to compute pixel-aligned features zero-shot from off-the-shelf foundation models (such as CLIP~\cite{clip}, AudioCLIP~\cite{audioclip}, and variants).

\subsection{Fusing pixel-aligned foundation features to 3D}
\label{sec:fusing_features}

\textbf{Map representation}: We represent our open-set multimodal 3D map $\mathcal{M}$ as an unordered set of points (indexed by $k$), each with the following attributes: (a) a vertex position $\overline{\boldsymbol{v}}_k \in \mathbb{R}^3$, (b) a normal vector $\overline{\boldsymbol{n}}_k \in \mathbb{R}^3$, (c) a confidence count $\bar{c}_k \in \mathbb{R}$, (d) a 3D color vector (optional), and (e) a \emph{concept vector} $\fs_k$ enabling open-ended querying.

\textbf{Frame preprocessing}: Each incoming frame $I_t$ (comprising a color image $C_t$ and a depth image $D_t$) is preprocessed to compute vertex-normal maps ($V_t$, $N_t$) and camera pose estimates $P_t$~\cite{pointfusionbase,park2017colored}.
Additionally, as described in Sec. \ref{sec:compute_features}, we compute the semantic context embedding $\fs_{u,v,t} \in \fs_{Xt}$ for each pixel in the input image $X_t$.

\begin{figure}[!t]
\includegraphics[width=\linewidth, trim={ 1.2cm 0.1cm 1.2cm 0},clip]{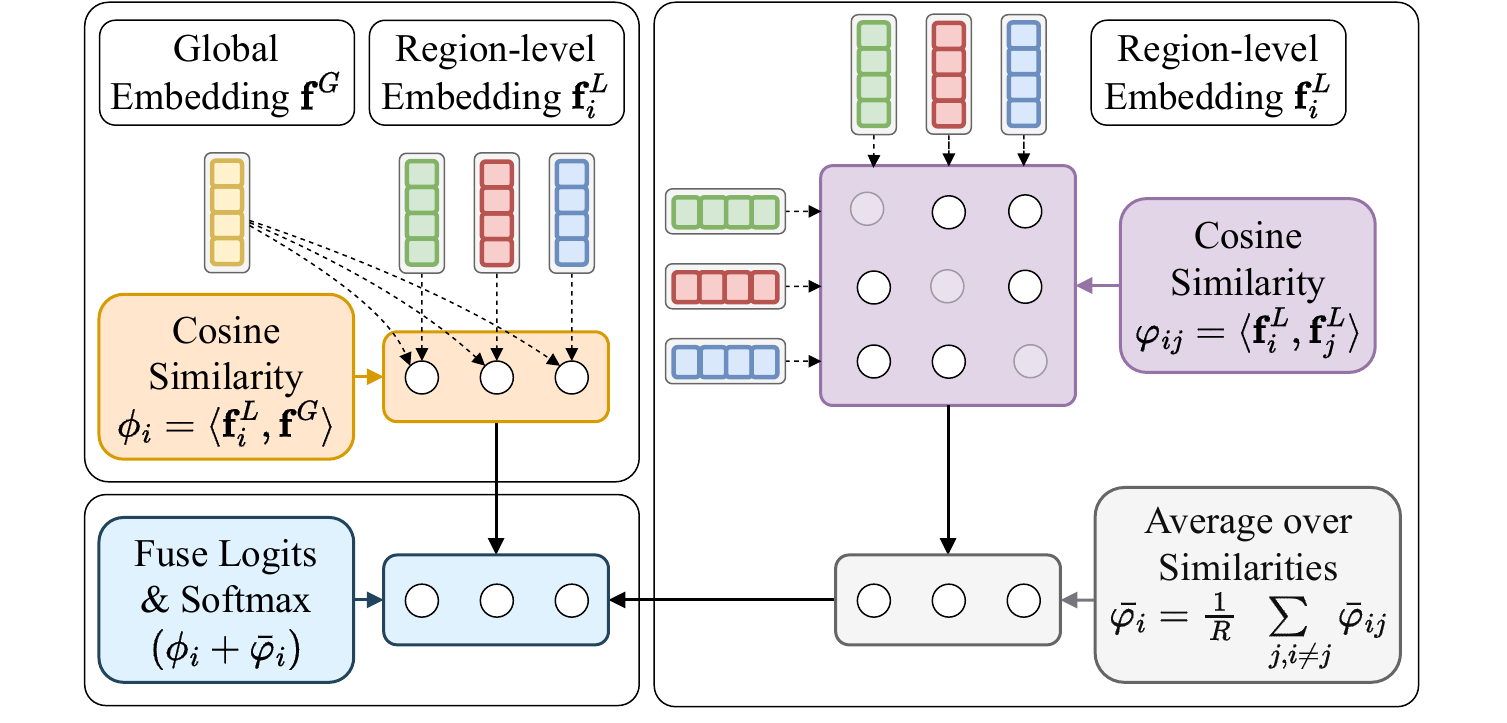}
\caption{For each image, the global ($\fg$) and local ($\fl$) features are fused to obtain our \textbf{pixel-aligned} features ($\fs$). \textbf{Top-Left:} We first compute cosine similarities between each local feature ($\fl$) with the global feature ($\fg$). \textbf{Right:} We compute an inter-feature similarity matrix, and compute the average similarity of each local feature to every other local feature, denoted $\bar{\varphi}_i$.
\textbf{Bottom-Left:} We combine these similarities to produce weights for fusing $\fg$ and $\fl$ to obtain pixel-aligned features $\fs$. See Sec. \ref{sec:compute_features} for details.}
\label{fig:fusionmath}
\end{figure}

\textbf{Feature fusion}: We fuse $\fs_{u,v,t}$ and $X_t$ into the global map following a 3D reconstruction pipeline~\cite{pointfusionbase}. 
First, vertex and normal maps are mapped to the global (map) coordinate frame using the camera pose $P_t$.
We then filter out points with noisy depth values by following the depth map fusion procedure outlined in ~\cite{pointfusionbase}.
The remaining points are fused into the global map $\mathcal{M}$.
A key departure from dense mapping approaches is the fusion of \emph{concept vectors}
$\fs_{u,v,t}$ in addition to depth (and optionally, color).
For each pixel $(u, v)_t$ in the image $X_t$ that have a corresponding point $p_k$ in $\mathcal{M}$, we integrate the features using the following scheme.
\begin{align}
    \fs_{k,t} &\leftarrow 
    \frac{\bar{c}_k\fs_{k,t-1} +  \alpha \fs_{u,v,t}}{\bar{c}_k + \alpha}\\
    \bar{c}_k &\leftarrow \bar{c}_k + \alpha
\end{align}
where $\alpha = e^{{-\gamma}^{2}/{2\sigma^2}}$ is the confidence assigned to each pixel-grounded feature assigned to the vertex being aggregated, $\gamma$ is the radial distance, and $\sigma=0.6$ is a scaling term.
We find empirically that having a confidence value based on the normalized radial distance to the camera center, similar to ~\cite{pointfusionbase, curless1996volumetric} works well.
We refer to the appendix for hyperparameter values and more details.

\subsection{Computing pixel-aligned features}
\label{sec:compute_features}

\begin{figure}[!t]
    \centering
    \includegraphics[width=\linewidth]{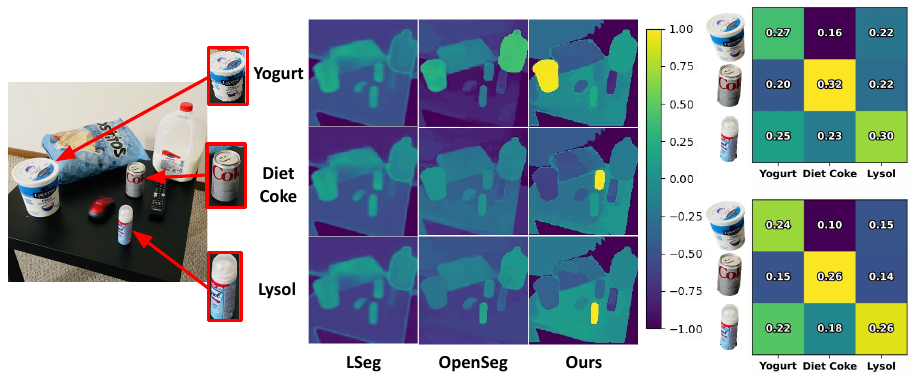}
    \caption{Our approach to computing \textbf{pixel-aligned features} is adept at capturing long-tailed and fine-grained concepts. The plots to the right show the similarity scores between the embeddings of the cropped image regions corresponding to \textit{diet coke}, \textit{lysol}, and \textit{yogurt} and their text embeddings, predicted by the base CLIP model used by LSeg and OpenSeg respectively. This implies that the base CLIP models know these concepts, yet, as can be seen on the tiled plots (center), LSeg and OpenSeg are not able to retrieve these concepts; they forget the concepts when finetuned. On the other hand, our zero-shot pixel-alignment approach does not suffer this drawback, and clearly delineates the corresponding pixels.}
    \label{fig:pixel-aligned-qualitative}
\end{figure}

Although some approaches have adapted foundation models like CLIP~\cite{lseg, openseg, clipseg}, these require additional training on labelled image-text data.
This results in pixel-alignment and stronger performance on in-dataset concepts, but we observe (see Fig.~\ref{fig:pixel-aligned-qualitative}) that the models struggle with fine-grained concepts absent in the finetuning datasets.
To mitigate this, we introduce a novel mechanism to construct pixel-aligned features that combine global (image-level) context encapsulated in models like CLIP, with local (region-level) information.

\textbf{Overview}: Fig.~\ref{fig:pipeline} gives us a broad overview of this section. Given an input image $ X \in \mathbb{R}^{3 \times H \times W}$, our method uses a foundation model $\mathcal{F}$ as a feature extractor to produce three types of embeddings, which we refer to as  global ($\fg$), pixel-aligned ($\fs$), and local ($\fl$).
First, the global embedding \mbox{$\fg=\mathcal{F}(X)$} is simply the embedding of the entire image.

\textbf{Local embeddings}: We employ a \textit{universal} instance segmentation model~\cite{cheng2021maskformer,sam} to produce a set of $R$ class-agnostic mask proposals (corresponding to image-regions that could potentially contain objects).
Next, for each region $r_i \in \mathbf{R}$, we define a bounding box $\textnormal{b}_{i} = \texttt{bbox}(r_i)$ around it.
The bounding boxes are fed through the model $\mathcal{F}$ to obtain local embeddings $ \fl_i = \mathcal{F}(\textnormal{b}_i)$.

\textbf{Fusing local and global features}: The pixel-aligned embedding for each region is a weighted combination of the global embedding and the corresponding local embedding.
The weight for each local embedding is based on its similarity with the global embedding $\fg$, as well as with other embeddings $\fl_i$.
Concretely, we compute the cosine similarity
\begin{equation}
    \phi_{i} = \left\langle \fl_i, \fg \right\rangle = \frac{(\fl_i)^T \fg}{\|\fl_i\| \|\fg\| + \epsilon}  
\end{equation}
between the local feature $\fl_i$ and the global feature $\fg_i$; and a matrix of cosine similarities between all pairs of local embeddings as $\varphi_{ij} = \langle \fl_i, \fl_j \rangle : \forall i, j \;  \text{such that} \; r_i, r_j \in \mathbf{R}$.
Next, for each local embedding $\fl_i$, we compute its average similarity to all other local embeddings $\bar{\varphi}_i$.
This accounts for the \emph{uniqueness} of the region for $\fl_i$ in the image.
\begin{equation}
    \bar{\varphi}_{i} =  \frac{1}{R}  \sum_{j=1, j\neq i}^{R} \varphi_{ij}
\label{eq:uniqueness}
\end{equation}
We combine the two similarities above to compute the mixing weight $w_{i} \in [0, 1]$ (with a temperature $\tau$, set to $1$ in all reported results).
\begin{equation}
    w_i = \frac{
    \displaystyle\exp{\left( \frac{
    \phi_i + \bar{\varphi}_{i} }{
    \tau} \right)}}{
    \sum_{i=1}^{R}
    \displaystyle\exp{\left( \frac{
    \phi_i + \bar{\varphi}_{i}}{
    \tau} \right)} }
\end{equation}
Finally, the pixel-aligned feature for each region $r_i$ is
\begin{equation}
    \fs_i = w_i \fg + (1- w_i) \fl
\end{equation}
which is normalized and mapped to the pixels $(u,v)$ in $r_i$.
We allow each pixel to belong to multiple regions; the corresponding pixel-aligned embedding $\fs_{u,v}$ is normalized once it accumulates features $\fs_i$ from regions $r_i$.

\begin{figure}[!t]
    \centering
    \includegraphics[width=.9\linewidth]{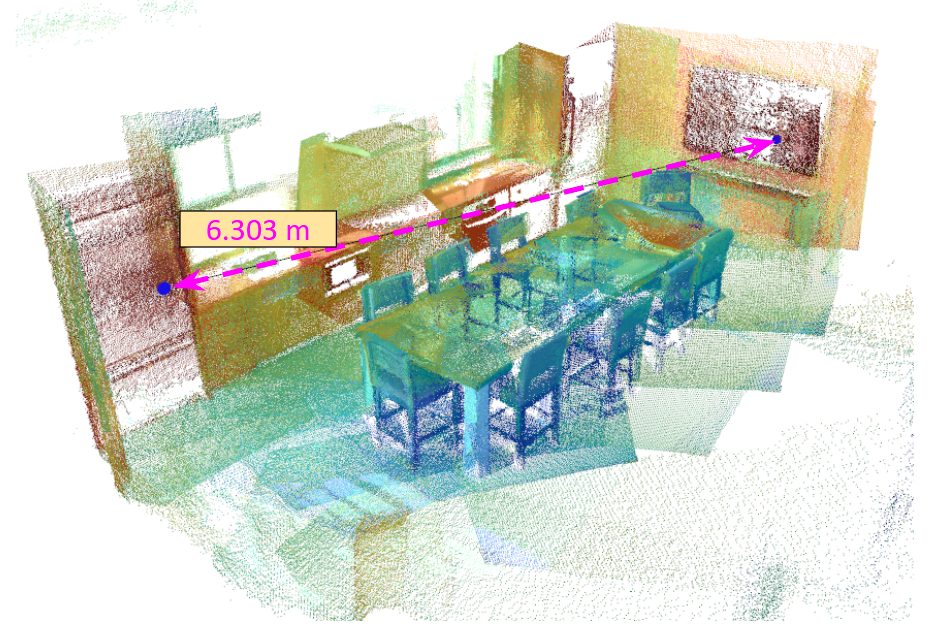}
    \caption{\textbf{3D spatial reasoning abilities}: A key benefit of lifting foundation features to 3D is the ability to reason about spatial attributes. For example, the query ``\textit{how far is the refrigerator from the television}''. gets parsed by our LLM-parser into the 3DSC \texttt{howfar(refrigerator, television)}. The centroid of the point set returned by the query term \texttt{refrigerator} and \texttt{television} are shown as blue circles, and the estimated distance between them (6.303 metres) as a straight line.}
    \label{fig:3d-spatial-reasoning}
\end{figure}

\textbf{Capturing long-tailed concepts}: We find that our pixel-aligned embeddings capture fine-grained and long-tailed concepts significantly better than approaches like LSeg~\cite{lseg} and OpenSeg~\cite{openseg}, which align CLIP features to pixels by training over smaller labelled datasets.
We illustrate this in Fig.~\ref{fig:pixel-aligned-qualitative}. We observe (right panel) that the underlying backbone CLIP models used by both LSeg and CLIP know the concepts \emph{diet coke}, \emph{lysol}, and \emph{yogurt}; however the finetuned (pixel-aligned) models do not.
This is due to forgetting phenomenon when finetuning CLIP-like models, as corroborated in~\cite{dont-stop-learning-clip}. 
LSeg and OpenSeg need to be finetuned on datasets with limited concepts, in order to obtain the ability of segmentation. However, this finetuning process harms their zero-shot ability to generalize to long-tailed and fine-grained concepts. 
In contrast, \coolname{} presents a new way of mapping foundation features to pixels and 3D points, and therefore it remains zero-shot and accurately aligns long-tailed concepts to the corresponding pixels, as shown by the middle plots in Fig.~\ref{fig:pixel-aligned-qualitative}.

\subsection{Multimodal querying over 3D feature-fused maps}
\label{sec:approach-querying}

\begin{figure}[!t]
\centering
\includegraphics[width=\linewidth]{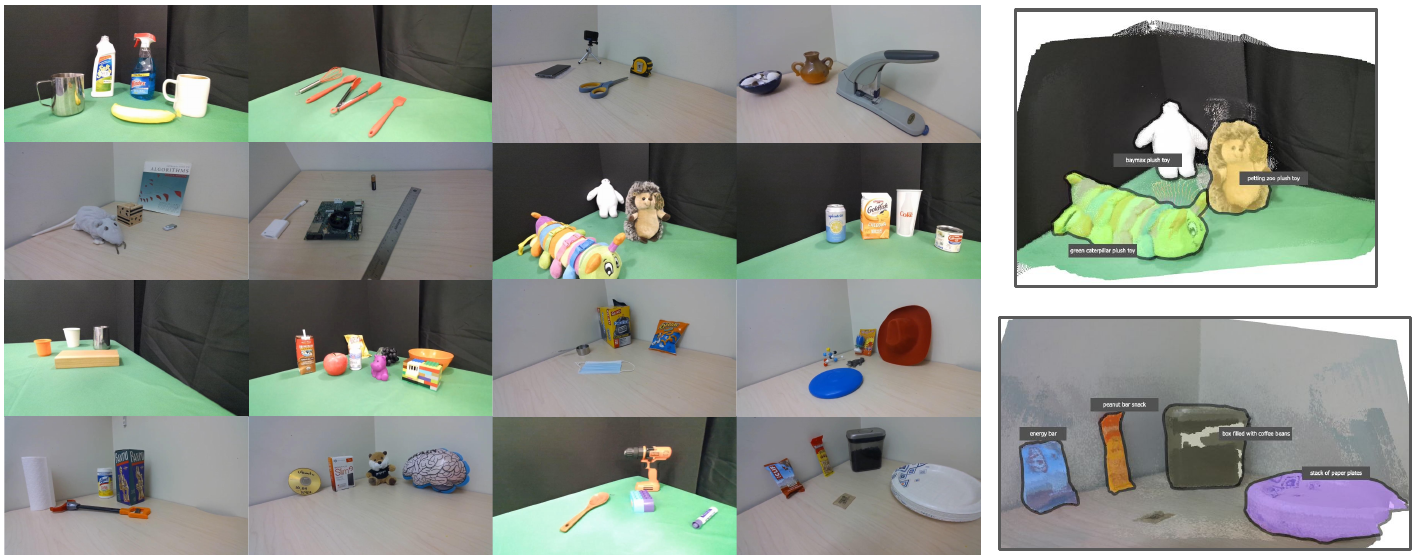}
\caption{Sample sequences from the \textbf{UnCoCo dataset} we captured to evaluate long-tailed reasoning over open-set multimodal 3D maps. To the right, we show sample reconstructions and semantic annotations over two sub-sequences. For each scene, UnCoCo provides 2D and 3D segmentation masks, and text, image, click, and audio queries.}
\label{fig:uncoco-samples}
\end{figure}

The 3D maps reconstructed by \coolname{} can be queried using multiple modalities. Given a query $q_{\text{mode}}$, and a map with fused concepts $\fs_k$, we compute a per-point score $s_k \in \left[-1, 1\right]$ as the cosine similarity defined as $ s_k = \left\langle f_k, q_{\text{mode}}\right\rangle$.
The computation of $q_{\text{mode}}$ changes depending on the modality used for querying.
\begin{enumerate}%
    \item \textbf{Text query}: $q_{\text{text}}$ is computed using the corresponding CLIP text encoder $\mathcal{F}_{\text{text}}$.
    \item \textbf{Click query}: $q_{\text{click}}$ is taken to be the fused feature vector $\fs_k$ at the clicked point.
    \item \textbf{Image query}: $q_{\text{image}}$ is computed as the image-level CLIP embedding of the query image.
    \item \textbf{Audio query}: $q_{\text{audio}}$ is the AudioCLIP~\cite{audioclip} embedding of the query sound clip.
\end{enumerate}

The similarities can then be post-processed by thresholding based on score, non-maxima suppression, and (optionally) clustering to produce 3D regions of interest.

\subsection{Building complex 3D spatial query modules}
\label{sec:spatial-query-building-blocks}

\begin{table*}[!th]
\centering
\begin{minipage}{.49\linewidth}
    \adjustbox{max width=\linewidth}{%
    \begin{tabular}{llcccc}
    \toprule
            & & 3D mIoU        & IoU \textgreater 0.15 & IoU \textgreater 0.25 & IoU \textgreater 0.5 \\
    \midrule
    \rowcolor{gray!20} & LSeg-3D    & 0.128          & 25\%                  & 16.66\%               & 9.72\%               \\
    \rowcolor{gray!20} \multirow{-2}{*}{Supervised} & OpenSeg-3D & 0.289          & 43.05\%               & 36.11\%               & 27.78\%              \\
    & MaskCLIP-3D & 0.091          & 25.97\%               & 9.09\%               & 1.30\%              \\
    & \bf \coolname{}    & \textbf{0.446} & \textbf{77.78\%}      & \textbf{69.44\%}      & \textbf{45.83\%}  \\
    \bottomrule 
    \end{tabular}
    } %
    \caption{{\protect Text-query based object localization performance on UnCoCo -- the \textit{structured} subset. In each column, a higher value corresponds to superior performance.}}
    \label{table:textquery-structured}
\end{minipage}\hfill
\begin{minipage}{0.49\linewidth}
    \adjustbox{max width=\linewidth}{%
    \begin{tabular}{llcccc}
    \toprule
     & & 3D mIoU        & IoU \textgreater 0.15 & IoU \textgreater 0.25 & IoU \textgreater 0.5 \\
    \midrule
     \rowcolor{gray!20}  & LSeg-3D     & 0.122          & 31.45\%               & 20.65\%               & 5.65\%               \\
    \rowcolor{gray!20} \multirow{-2}{*}{Supervised} & OpenSeg-3D   & 0.153          & 27.26\%               & 21.94\%               & 11.29\%              \\
    \multirow{2}{*}{Zero-Shot} & MaskCLIP-3D & 0.092          & 20.63\%               & 11.88\%               & 3.06\%              \\
    & \bf \coolname{}      & \textbf{0.378} & \textbf{70.16\%}      & \textbf{59.52\%}      & \textbf{34.03\%}    \\
    \bottomrule 
    \end{tabular}
    } %
    \caption{Text-query based detection performance on UnCoCo -- the \textit{unstructured} subset. Results averaged over 20 trials. In each column, a higher value corresponds to superior performance.
    }
    \label{table:textquery-unstructured}
\end{minipage}
\begin{minipage}{0.49\linewidth}
    \adjustbox{max width=\textwidth}{%
    \begin{tabular}{llcccc}
    \toprule
     &  & 3D mIoU        & IoU \textgreater 0.15 & IoU \textgreater 0.25 & IoU \textgreater 0.5 \\
    \midrule
    \rowcolor{gray!20}  & LSeg-3D               & 0.134          & 26.88\%               & 21.51\%               & 9.68\%               \\
    \rowcolor{gray!20} \multirow{-2}{*}{Supervised} & OpenSeg-3D             & 0.112          & 23.66\%               & 18.28\%               & 8.60\%               \\
    \multirow{2}{*}{Zero-Shot} & MaskCLIP-3D & 0.094          & 21.51\%               & 11.83\%               & 4.30\%              \\
    & \bf \coolname{}                & \textbf{0.331} & \textbf{54.84\%}      & \textbf{51.61\%}      & \textbf{31.18\%}    \\
    \bottomrule 
    \end{tabular}
    } %
    \caption{Image-query based detection performance on UnCoCo -- the \textit{structured} subset. Results averaged over 3 trials.}
    \label{table:imagequery-structured}
\end{minipage}\hfill
\begin{minipage}{0.49\linewidth}
    \centering
    \adjustbox{max width=.8\textwidth}{%
    \begin{tabular}{llcc}
    \toprule
      & &  Accuracy (\%)  &  IoU \\
    \midrule
    \multirow{3}{*}{source-ambiguous} & Random & 7.14\%  & N/A \\ & AudioCLIP~\cite{audioclip} & 23.81\% & N/A \\ & \textbf{\coolname{}} & \textbf{64.29\%} & 0.287 \\
    \midrule
    \multirow{3}{*}{ecological} & Random & 5.56\%  & N/A \\ & AudioCLIP~\cite{audioclip} & 22.22\% & N/A \\ & \textbf{\coolname{}}  & \textbf{66.67\%} & 0.301 \\
    \bottomrule
    \end{tabular}
    } %
    \caption{Audio-query based detection and classification performance on UnCoCo.}
    \label{table:audioquery}
\end{minipage}
\vspace{-0.2cm}
\end{table*}

Unique capabilities unlocked by fusing features into 3D space include the ability to reason about objects that were never co-observed in an image, and the ability to reason about spatial attributes (such as relative positions, orientations, support, containment, etc.) accessible only from 3D representations.
To this end, we leverage the computed similarity scores to build a set of \emph{3D spatial comparator} (3DSC) modules that may be further composed to recover more complex attributes.

Our set of 3DSCs all take on the relation signature \textsc{relation(query$_a$, query$_b$)} and return a scalar or boolean value as appropriate. The complete set of 3DSCs includes (see supplementary material for more details)
\begin{enumerate}
    \item The \textsc{HowFar($q_a$, $q_b$)} 3DSC returns the distance of objects referenced by queries $q_a$ and $q_b$ respectively.
    \item The boolean 3DSCs \textsc{IsToTheRight($q_a$, $q_b$)}, \textsc{IsToTheLeft($q_a$, $q_b$)}, \textsc{OnTopOf($q_a$, $q_b$)}, \textsc{Under($q_a$, $q_b$)} return \textsc{true} or \textsc{false} depending on whether or not the object referenced by queries $q_a$ and $q_b$ satisfy the appropriate spatial relationship (relative to a specified viewing direction).
\end{enumerate}
In Sec.~\ref{sec:outlook}, we optionally adopt a large language model~\cite{gpt3} for parsing language queries to an appropriate composition of 3DSCs.
For instance, the query ``\emph{what is the distance between the refrigerator and the television?}'' is parsed into \textsc{howFar(\textit{refrigerator}, \textit{television})}. This is depicted in Fig.~\ref{fig:3d-spatial-reasoning}.

However, for all other results presented in this paper---unless otherwise specified---the language queries are directly fed into the CLIP text encoder without any preprocessing.

\textbf{Implementation details}: Our feature fusion algorithm is implemented on top of the $\nabla SLAM$~\cite{gradslam} dense SLAM system, as this was one of the few implementations of the PointFusion algorithm~\cite{pointfusionbase}, and for its convenience of interfacing with PyTorch for computing and accessing foundation features.
For generating class-agnostic (generic) object masks, we use the Mask2Former~\cite{cheng2021mask2former} or the segment anything (SAM)~\cite{sam} models for category-agnostic instance segmentation.
When using Mask2Former, we generate 100 mask proposals per image.
When using SAM, we find that there is significantly lower redundancy in the number of mask proposals -- we therefore drop our uniqueness term in Eq.~\ref{eq:uniqueness}.
Our odometry and mapping approaches run at frame-rate (15 Hz). The pixel-aligned feature extraction processes run offline (10–15 seconds / image) on an NVIDIA RTX 3090 GPU.

\textbf{Real-time inference}: To optimize the performance and efficiency of the foundation models employed (SAM~\cite{sam}, DINO~\cite{dino}, and CLIP~\cite{clip}), we use standard quantization and tracing methods.
By applying both quantization and tracing techniques to our models, we are able to achieve significant improvements in their efficiency, without compromising their accuracy.
This enables us to deploy our models in real-world scenarios, where memory and runtime efficiency are crucial.

\section{Case studies}

We design a systematic experimental study to investigate the following questions:
\begin{compactenum}
\item How do open-set multimodal 3D maps fare when queried using text, images, clicks, or audio?
\item How do we leverage the rich concept space embedded to 3D for spatial reasoning?
\item How well does \coolname{} work on real-world robotics tasks?
\item What previously infeasible downstream use-cases can \coolname{} enable?
\end{compactenum}

\textbf{Experimental setup}: Our experimental benchmark comprises of sequences from multiple publicly available datasets, and sequences we collect.
The benchmark comprises 20 indoor (apartment-scale) scenes from ScanNet~\cite{scannet, scannet200}, Replica~\cite{replica}, and self-captured sequences; 5 outdoor (urban driving) scenes; 20 indoor (tabletop) scenes with common household products (UnCoCo); and 5 synthetic scenes from the ICL~\cite{icldataset} and iTHOR benchmarks~\cite{ai2thor}.

\begin{figure*}[!t]
    \centering
    \includegraphics[width=\linewidth]{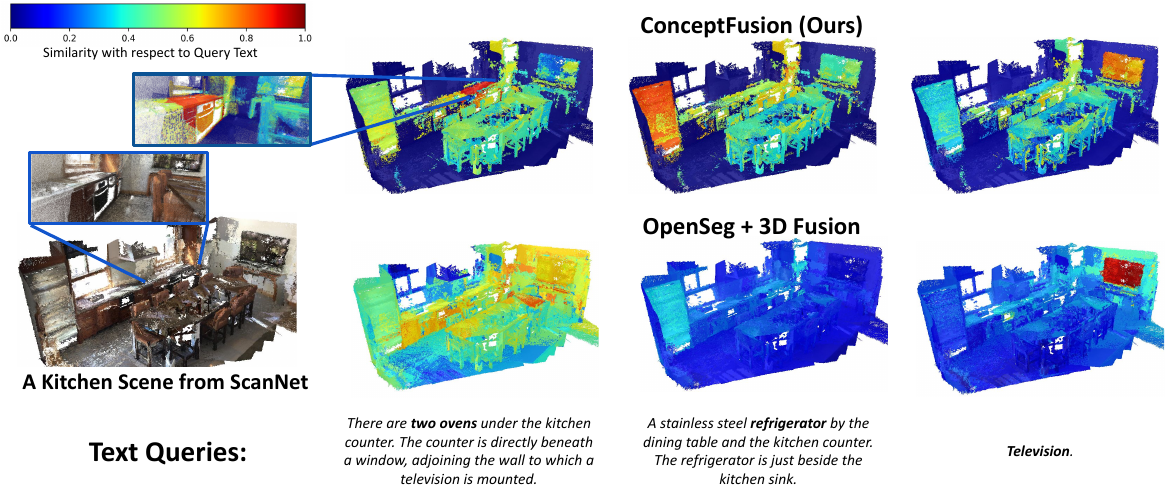}
    \caption{\textbf{Text queries over ScanNet~\cite{scannet}}: \coolname{} is able to handle long-form text queries and accurately localize objects referenced by the query. In the first two scenarios, OpenSeg~\cite{openseg} is distracted by the presence of several confounding attributes (\textit{kitchen counter}, \textit{window}, \textit{television}). The third scenario shows a single world query (\textit{television}) that is part of the COCO Captions \cite{cococaptions} dataset used to train OpenSeg, placing it at an advantage (and hence resulting in a more discriminable heatmap). \coolname{}, nonetheless, accurately assigns the highest response to the  map points representing the television. In each query, the referenced object is \textbf{boldfaced}.}
    \label{fig:scanrefer-qualitative}
\end{figure*}

\textbf{Approaches evaluated}:
Since there is no prior work on constructing open-set multimodal maps of the kind we build with \coolname{}, we make a best-effort comparison with concurrent work in this space.
Approaches such as VL-Maps~\cite{huang23vlmaps}, NLMap-SayCan~\cite{chen2022nlmapsaycan}, CoWs~\cite{gadre2022clip,gadre2022cow}, CLIP-Fields~\cite{clipfields} leverage LSeg~\cite{lseg}; while OpenScene~\cite{openscene} experiments with both LSeg~\cite{lseg} and OpenSeg~\cite{openseg}.
We therefore implement two baseline approaches that leverage LSeg and OpenSeg features respectively, and apply our feature fusion technique to obtain open-set 3D maps.
\emph{We refer to these baselines as \textbf{LSeg-3D} and \textbf{OpenSeg-3D} respectively}.
Additionally, to compare with a state-of-the-art zero-shot segmentation approach, we also implement \emph{\textbf{MaskCLIP-3D}}, which fuses per-pixel MaskCLIP~\cite{maskclip} features into a 3D map.

\subsection{Multimodal queries on the UnCoCo dataset}
\label{sec:uncoco-eval}

To the best of our knowledge, there is no existing system that supports queries as diverse and in as many modes as \coolname{}; and consequently, there are no publicly available dataset to evaluate such a system on.

\textbf{UnCoCo}: To overcome this gap, we captured our own dataset; which we call \textit{UnCoCo}, short for \textit{Uncommon contexts for Common Concepts}. This real-world dataset comprises 3D scans of 78 commonly found household and office objects on a tabletop surface (see Fig.~\ref{fig:uncoco-samples}).
There are 20 RGB-D image sequences in all, comprising a total of 12075 color and depth image pairs. For each image, we provide per-object 2D instance segmentation masks, and for each scene, a corresponding 3D segmentation mask.
Importantly, UnCoCo supports a diverse set of query modalities -- text, click, image, and audio.
For each of these query modalities, we provide a \textit{structured} and an \textit{unstructured} set.
The structured set comprises carefully curated and vetted queries, whereas the (much larger) unstructured set scrapes query text, and images from web-scale data.
Each image in this dataset has 3-5 objects; each object has one structured text query, and 5-40 unstructured text queries (freeform queries, crowdsourced from human annotators); up to 20 structured click queries and up to 100-2000 unstructured click queries; 10 structured image queries and 10-100 unstructured image queries; and 1-5 structured audio queries.
With a little over 500000 queries across modalities, UnCoCo is the only dataset to date supporting multimodal 3D query evaluations on commonly found objects.

\textbf{Text query}: We evaluate text-query based object localization performance on 3D maps, on the UnCoCo dataset. This task is extremely challenging due to the versatility of objects present in the dataset, ranging from extremely small objects (e.g., a 4-gram sachet of sugar, whiteboard markers), to thin objects (e.g., face masks, compact discs), to nonconvex geometries (e.g., a whisk, lego block constructions, shells).
We evaluate two state-of-the-art per-pixel CLIP-aligned feature extractors in LSeg~\cite{lseg} and OpenSeg~\cite{openseg}, which require additional training over a large labelled dataset; and MaskCLIP~\cite{maskclip} -- the current state-of-the-art approach for extracting zero-shot per-pixel labels based on a text prompt.
Results are shown in Table~\ref{table:textquery-structured}. For each technique evaluated, we report the 3D mean intersection-over-union (IoU) metric, and also detection accuracies at IoU thresholds of 0.15, 0.25, and 0.5.
We see that \coolname{} outperforms all other approaches by a significant margin. We attribute this to two key characteristics of \coolname{}. First, \coolname{} operates on the unmodified CLIP feature space, whereas approaches like LSeg and OpenSeg specialize to the datasets they are finetuned on and end up gradually forgetting concepts that are infrequent on the finetuning set.
Second, \coolname{} features efficiently combine global (image-level) features with local (region-level) context; providing a rich pixel-level (and subsequently point-level) grounding. We also observe a similar trend over the unstructured text query set, as reported in Table~\ref{table:textquery-unstructured}. \coolname{} exhibits more graceful performance degradation to unstructured queries (long sentences).

\textbf{Image query}: In Table~\ref{table:imagequery-structured}, we report results for the scenario where the query concept is presented in the form of an image. For instance, the goal of finding a \textit{can of soda} in a 3D scene is specified by providing a randomly picked image of a soda can from the web. Here, we again observe that \coolname{} outperforms other finetuned foundation models by a significant margin in terms of both 3D mIoU and detection accuracy. Interestingly, we note that performances of most approaches across the text and image modalities remain consistent; as evident from the mIoU and detection metrics.

\textbf{Audio query}: A unique capability of our approach is to localize 3D objects based on audio queries. We conduct experiments on the UnCoCo dataset, which contains two classes of audio queries. \textit{Source-ambiguous} queries comprise sounds that are caused due to material properties and geometries of objects and commonly involve object motions including scratching, scraping, rolling, crushing, tearing, etc.
\textit{Ecological} queries comprise sounds that are unique to an object (or category) -- these include sounds like spraying, drilling, stirring, knocking, clicking, etc.
To compensate for the lack of open-set sound source localization baselines, we implement a privileged-information baseline by providing AudioCLIP~\cite{audioclip} with a set of ground-truth instance boxes per image; and using the computed similarity scores to rank the relevance of each box to the query audio.
Results are presented in Table~\ref{table:audioquery}.
However, we notice that, using purely local context (as with the AudioCLIP baseline) is not performant enough; and that \coolname{} features, which fuse global and local contexts, again perform better than purely-local features.

\subsection{Open-Set semantic segmentation on other datasets}

We also evaluate semantic segmentation performance on existing datasets with well-known concepts (classes either directly or indirectly accessible via COCO Captions~\cite{cococaptions}, a dataset used for finetuning by other pixel-aligned models).
In particular, we provide results on validation subsets of the ScanNet~\cite{scannet}, Replica~\cite{replica}, and the SemanticKITTI~\cite{semantickitti} datasets and present results in Table~\ref{table:textquery-scannet-2d} (refer to our appendix for further details).
Of the approaches presented here, LSeg requires per-pixel CLIP features as labels, OpenSeg leverages per-image captions for labels, and CLIPSeg trains a shallow decoder atop the CLIP image encoder.
MaskCLIP is the closest zero-shot baseline; we outperform it by a large margin.

\begin{figure}[!t]
    \centering
    \includegraphics[width=\linewidth]{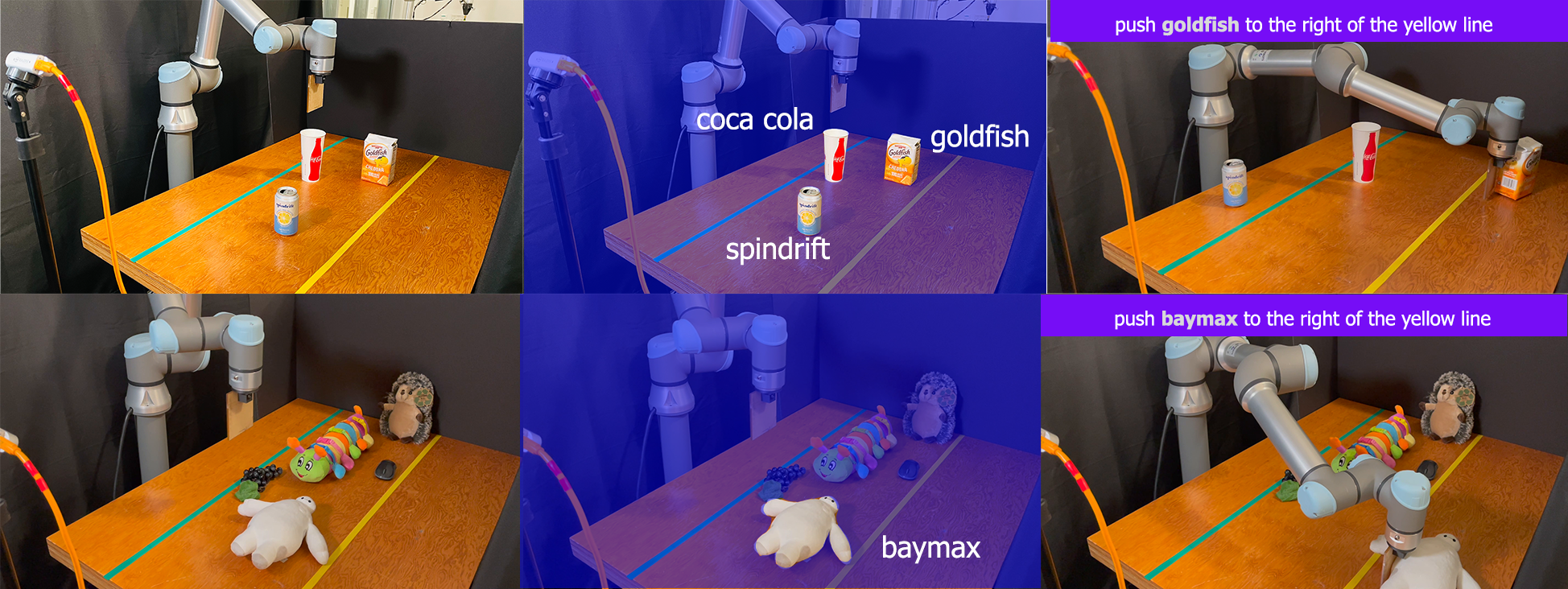}
    \caption{Real-world \textbf{tabletop rearrangement} experiments. The robot is provided with rearrangment goals involving novel objects. (\textit{Top row}) \textit{push goldfish to the right of the yellow line}, where \textit{goldfish} refers to the brandname of the pack of Cheddar snack. (\textit{Bottom row}) \textit{push baymax to the right of the yellow line}, where \textit{baymax} refers to the plush toy depicting the famous Disney character.
    }
    \label{fig:real-world-rearrangement}
\end{figure}

\begin{table}[!t]
\centering
\caption{\textbf{Open-set semantic segmentation} on ScanNet~\cite{scannet}, Replica~\cite{replica}, and SemanticKITTI~\cite{semantickitti}. Queries provided in the form of text labels. \textbf{Priv.} (shaded area) indicates privileged approaches where off-the-shelf CLIP models are finetuned specifically for semantic segmentation. \textbf{ZS} (unshaded area) evaluates zero-shot approaches. We outperform the closest zero-shot approach, MaskCLIP~\cite{maskclip} by a large margin across datasets. Furthermore, \coolname{} is competitive to privileged baselines for this task.}
\adjustbox{max width=.95\linewidth}{%
\begin{tabular}{llcccccc}
\toprule
& & \multicolumn{2}{c}{ScanNet} & \multicolumn{2}{c}{Replica} & \multicolumn{2}{c}{Semantic KITTI} \\
& & mAcc & f-mIoU & mAcc & f-mIoU  & mAcc & f-mIoU \\
\midrule
\rowcolor{gray!20} & LSeg~\cite{lseg} & 0.70 & 0.63 & 0.52 & 0.33 & 0.84 & 082      \\
\rowcolor{gray!20} & OpenSeg~\cite{openseg} & 0.63 & 0.62 & 0.54 & 0.41 & 0.78 &  0.77   \\
\rowcolor{gray!20} & CLIPSeg (rd64-uni)~\cite{clipseg} & 0.41 & 0.34 & 0.32 & 0.23 & 0.77 &  0.75   \\
\rowcolor{gray!20} & CLIPSeg (rd16-uni)~\cite{clipseg} & 0.41 & 0.36 & 0.40 & 0.28 & 0.79  &  0.77       \\
\rowcolor{gray!20} \multirow{-5}{*}{Priv.} & CLIPSeg (rd64-uni-refined)~\cite{clipseg}     & 0.23 & 0.24 & 0.13 & 0.13     &  0.28  &  0.26       \\
\multirow{3}{*}{ZS} & MaskCLIP~\cite{maskclip}    & 0.24 & 0.28 & 0.01 & 0.05    &  0.70 &  0.66        \\
& Mask2former + Global CLIP feat & 0.35 & 0.48 & 0.13 & 0.10 & 0.22	 & 0.20    \\
& \bf \coolname{} & \textbf{0.63} & \textbf{0.58} & \textbf{0.31} & \textbf{0.24}    &  \textbf{0.79} &  \textbf{0.78}         \\
\bottomrule
\end{tabular}
} %
\label{table:textquery-scannet-2d}
\end{table}

\begin{figure}[!t]
    \centering
    \includegraphics[width=.9\linewidth]{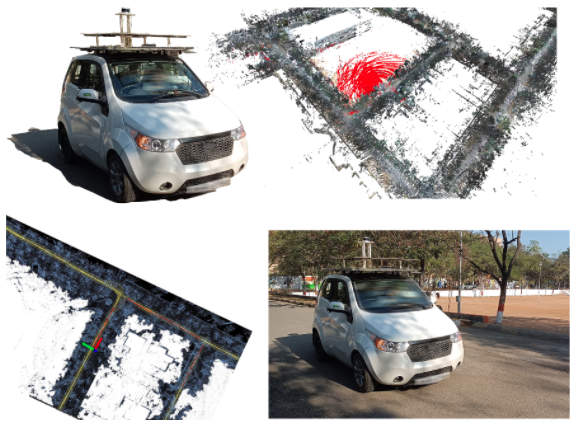}
    \caption{Real-world \textbf{autonomous navigation} experiments. (Left to right; top to bottom) Autonomous drive-by-wire platform deployed; pointcloud map  of the environment with the response to the openset text-query "football field" (shown in red); path found to the football field (shown in red); car successfully navigates to the destination autonomously. See our webpage for more results.}
    \label{fig:real-world-car}
\end{figure}

\begin{figure*}[!h]
    \centering
    \includegraphics[width=1.\linewidth]{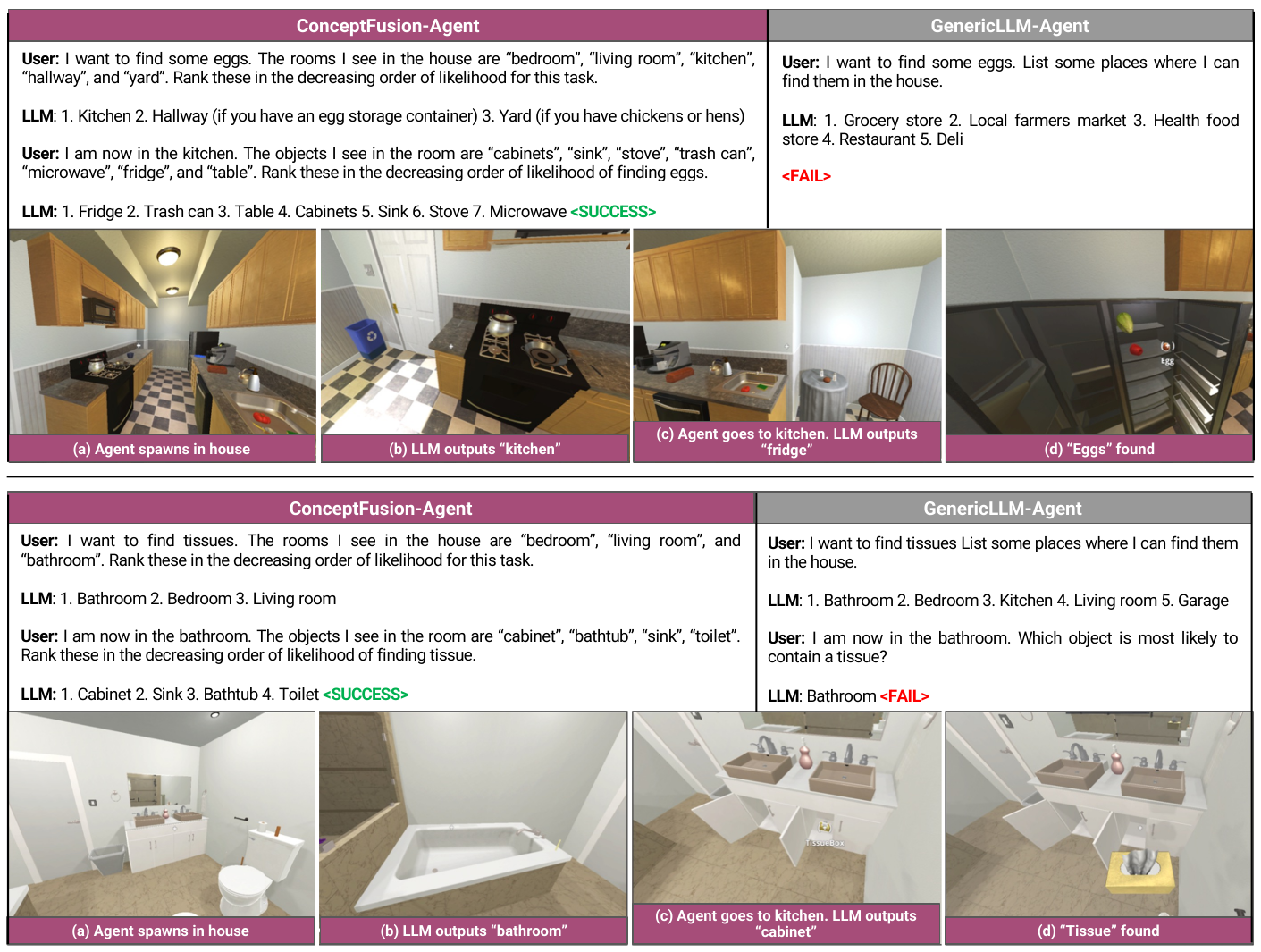}    \caption{\textbf{Integration of a large-language model (LLM) based planner in-the-loop}. We illustrate two scenarios from the AI2-THOR~\cite{ai2thor} interactive household simulator. The GenericLLM-Agent fails to achieve the specified task since it does not have an explicit 3D map representation, devoiding the LLM of the requisite context to accomplish the task. Our \coolname{}-Agent modifies its response by querying the open-set map to include rooms and object locations in its immediate vicinity. The LLM receives this context to guide the agent towards success.}
    \label{fig:gpt-integration}
\end{figure*}

\subsection{Experiments on real robotic systems}
\label{sec:real-robot-expts}

Our experiments on real robotic systems may also be accessed in a video format on our \href{\webpage}{companion website}.

\textbf{Zero-shot tabletop rearrangement}: To evaluate the applicability of \coolname{} to real-world robotic interaction, we conduct experiments on a zero-shot tabletop rearrangement task with a UR5e manipulator and an Intel Realsense D415 RGB-D camera.
The task involves a workspace (here a tabletop) with a few previously unseen objects in it. In some trials, the object set also includes distractors placed to hamper perception and/or manipulation planning.
Two sides of the workspace (see Fig.~\ref{fig:real-world-rearrangement}) are tagged \textit{left} and \textit{right} respectively (areas on either side of the table, as indicated by the green and yellow lines).
For each set of objects, a goal instruction is specified in the form of a natural language command. For instance, the two scenarios in Fig.~\ref{fig:real-world-rearrangement} correspond to the commands \textit{spindrift to the left; goldfish to the right; coca cola to the left} (top row) and \textit{baymax to the right} (bottom row).
This task challenges the perception capabilities of our system in multiple ways. First, \coolname{} must identify the correct object corresponding to the text query. Second, the object localization must be precise, to allow the end effector to execute a successful motion plan; moving the object to its goal location, while accounting for distractors.
After \coolname{} identifies and segments the object of interest, we estimate the object centroid as the median of the region of interest obtained over the depth image.
The robot arm then computes a motion plan (using the AIRobot library~\cite{airobot2019}) to push the object to the specified target region (i.e., to the \emph{left} or \emph{right}).

\textbf{Text-driven autonomous navigation}: We employ \coolname{} in open-set autonomous navigation for a self-driving platform.
Given a feature-fused map of an environment, we search the map for a best-match destination to a text query, and navigate autonomously to the location thus chosen.
We used a drive-by-wire autonomous vehicle equipped with a calibrated stereo camera and lidar to reconstruct a map of a 320000 square yard (4000 sq. m.) urban area.
The 3D map points are used by a lidar odometry algorithm (LeGO-LOAM~\cite{shan2018lego}) for global localization, and the fused features are used for open vocabulary queries to generate global waypoints.
Other components of the autonomy stack include a shortest-path global planner, a Frenet~\cite{frenet_planner} local planner for obstacle avoidance, and a Stanley-Controller~\cite{stanleycontroller} for trajectory tracking.
This allows us to navigate to open-set goals specified as natural language queries. In our field trials, we evaluated multiple goals, such as \emph{garbage bin}, \emph{roundabout}, \textit{entrance gate}. Fig.~\ref{fig:real-world-car} illustrates this process over a text query \emph{football field}. Please refer to our project page and supplementary video for viewing this demo.

\begin{table}[!h]
\centering
\caption{\textbf{Ablation analysis}: We ablate each component of our pixel-aligned feature fusion method. Fusing the image-level CLIP features with region-level features is key to good performance; while the uniqueness term contributes by suppressing rendundant mask proposals. The full system achieves the best performance.}
\label{table:ablation}
\begin{tabular}{lll}
\toprule
\multicolumn{1}{l}{\textbf{Approach}}               & \multicolumn{1}{l}{\textbf{mAcc}} & \multicolumn{1}{l}{\textbf{f-mIoU}} \\ \midrule
\multicolumn{1}{l}{Use global CLIP for every pixel} & \multicolumn{1}{l}{0.35}          & \multicolumn{1}{l}{0.48}            \\ 
\multicolumn{1}{l}{Mask2Former + Local CLIP}        & \multicolumn{1}{l}{0.43}          & \multicolumn{1}{l}{0.33}            \\ 
Remove uniqueness term (Eq. 4)                        & 0.55                               & 0.46                                 \\
\textbf{Ours (full)}                                  & \textbf{0.63}                      & \textbf{0.58}                 \\ \bottomrule      
\end{tabular}
\end{table}

\subsection{Ablation analyses}

\textbf{Pixel-alignment design choices}: We evaluate the design choices made in our pixel-alignment scheme on the ScanNet dataset. Results are presented in Table~\ref{table:ablation}. For each approach we indicate the mAcc (mean accuracy) and f-mIoU (frequency-weighted mean intersection-over-union). We see that, each component of the proposed method results in clear, significant improvement in performance. The first baseline (``use global CLIP") only use the global (image-level) CLIP embedding for each pixel in the image (a trivial baseline). The ``Mask2former + Local CLIP" variant takes in mask proposals from Mask2Former (i.e., the exact same proposals used in our full pipeline), but extracts features from each mask independently (i.e., does not apply equations 3 through 6). The ``Remove uniqueness term..." variant fuses features computed from individual masks with those computed over the entire image, but does not account for mask uniqueness (i.e., we skip equation~\ref{eq:uniqueness}). The ``Ours (full)" variant uses the full pipeline.

\textbf{3D vs 2.5D}: In Table~\ref{table:3d-spatial-reasoning}, we compare the performance of the full \coolname{} system with respect to a baseline system that uses only a single RGB-D image (i.e., a 2.5D image).
While the full details of the 3D spatial reasoning tasks are deferred to Sec.~\ref{sec:outlook}, we see that the 2.5D approach performs poorly on 3D reasoning tasks that require scene-level context typically absent in a single image.

\textbf{Choice of segmentation model}: We also assess the impact of the choice of the \emph{universal} segmentation model used in our system. We find that using the segment anything model (SAM)~\cite{sam} to produce class-agnostic instance masks results in significant performance boosts over using Mask2Former~\cite{cheng2021mask2former}. The results are presented in Table~\ref{tab:sam-vs-mask2former}

\section{Outlook}
\label{sec:outlook}

In this section, we provide perspectives on other emerging directions in foundation models for robotics, and how open-set multimodal 3D maps can augment and expand the capabilities of these models.

\begin{table}[!ht]
\centering
\caption{3D lifting enables \coolname{} to respond to spatial reasoning queries accurately. Each entry showcases success rates over specific query types (25 queries per type).}
\adjustbox{max width=\linewidth}{%
\begin{tabular}{lcccc}
    \toprule
     & Distance & Rel. position & Support & Containment \\
    \midrule
    Random & 24\% & 36\% & 52\% & 44\% \\
    2.5D (RGB-D image only) & 32\% & 28\% & 76\% & 68\% \\
    \bf \coolname{}  & \textbf{84\%} & \textbf{76\%} & \textbf{96\%} & \textbf{72\%} \\
    \bottomrule
\end{tabular}
} %
\label{table:3d-spatial-reasoning}
\end{table}

\begin{table}[!ht]
    \centering
    \caption{Impact of segmentation model on the performance of \coolname. We see that the segment anything model (SAM)~\cite{sam} offers a substantial performance boost over Mask2Former~\cite{cheng2021mask2former}. Ablation performed on the Replica~\cite{replica} dataset.}
    \begin{tabular}{ccc}
        \toprule
        \textbf{Approach} & \textbf{mAcc} & \textbf{f-mIoU} \\
        \midrule
        \coolname{} (w/ Mask2Former~\cite{cheng2021mask2former}) & 24.16 & 31.31 \\
        \coolname{} (w/ SAM~\cite{sam}) & \textbf{31.53} & \textbf{38.70} \\
         \bottomrule
    \end{tabular}
    \label{tab:sam-vs-mask2former}
\end{table}

\textbf{3D spatial reasoning abilities}: We also evaluate the unique ability of our queryable maps to reason about object spatial relationships in 3D.
 We generate a set of 100 (natural language) spatial reasoning queries over 5 scenes from the ScanRefer validation set.
The queries are split into 4 sets (25 queries per set), each designed to evaluate a specific type of 3D spatial relationship -- distances, relative positions (e.g., \textit{to the left of}), support (e.g., \textit{on top of}), and containment.
These queries are chosen such that (a) each query references two objects, and (b) two objects are co-observed in a single image (candidates are proposed using a distance-based threshold; and validated using image segmentation labels) for the distance and relative position query sets.
The results are presented in Table~\ref{table:3d-spatial-reasoning}, and compared against a baseline approach that uses only the pointcloud obtained by backprojecting a single RGB-D image (2.5D).
This baseline performs poorly in the distance and relative position queries, where the queries require reasoning about objects that were never co-observed in image space.
In the \textit{support} and \textit{containment} queries, the baseline shows better performance, as these are relations that can largely be gleaned from image observations alone. However, this approach still fails for room-level containment queries of type \texttt{is <OBJ> in <ROOM>}); which require additional context.
Fig.~\ref{fig:3d-spatial-reasoning} showcases an example of a distance query between two objects never co-observed in 2D.

\begin{figure*}[!th]
    \centering
    \includegraphics[width=.95\linewidth,trim={ 1cm 3.5cm 0 4cm},clip]{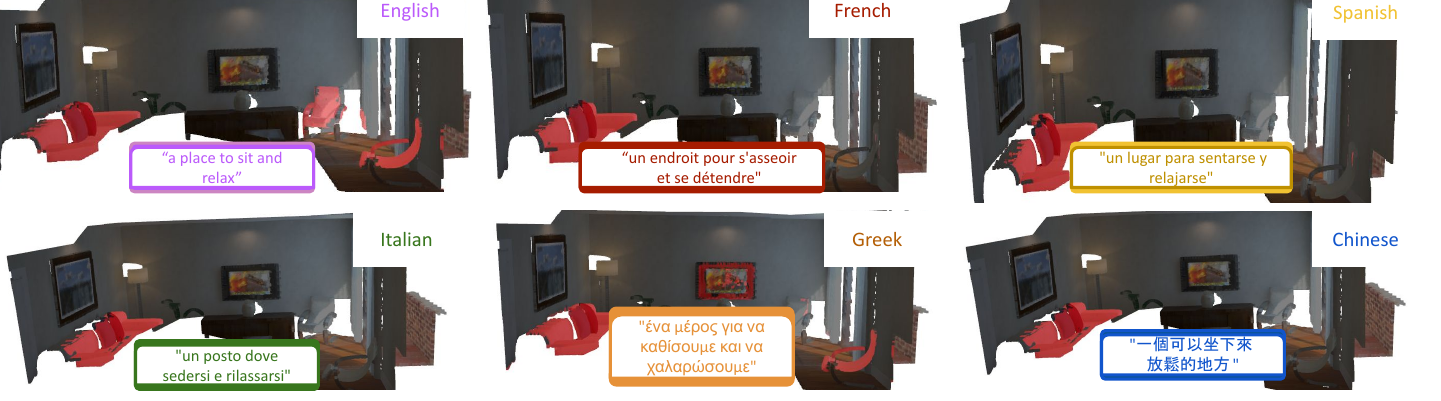}
    \caption{The zero-shot nature of our approach allows integration with newer off-the-shelf foundation models without the need for finetuning. We demonstrate this ability by leveraging multilingual CLIP models, which enables us to query the same concept across multiple languages.}
    \label{fig:multilingual}
\end{figure*}

\textbf{Integration with large language models (LLMs)}:
Systems such as SayCan~\cite{saycan2022arxiv} have demonstrated the viability of large-language models as queryable knowledge bases that may be used to generate task-level plans, to be executed by lower-level skills.
With \coolname{}, we can complement these task planning and reasoning abilities with a \emph{perception system} that can interface via natural language (and other modalities).
To demonstrate this, Fig.~\ref{fig:gpt-integration} illustrates two scenarios from the AI2-THOR~\cite{ai2thor} interactive household simulator.
In each scenario, the robot is equipped with the task of finding an object of interest \emph{that is not in its map}, because it is concealed within a receptacle.
We investigate the behavior of an \emph{GenericLLM-Agent} (a large language model~\cite{gpt3}) and a \emph{\coolname{}-Agent} (GenericLLM-Agent equipped with a set of rooms and objects queried from our map).
As seen in Fig.~\ref{fig:gpt-integration}, while the GenericLLM-Agent is able to generate seemingly plausible subgoals to achieve the task, the lack of knowledge of the map inhibits its success.
On the other hand, the \coolname{}-agent successfully achieves both tasks by restricting the generative capabilities of the LLM to scene-specific contexts.

\textbf{Multilingual abilities}:
Since \coolname{} is zero-shot, it is directly applicable to any existing image-level foundation model, without requiring additional finetuning.
To demonstrate this, we compute pixel-aligned features from multilingual CLIP models, and this extends the capabilities of our system to interpret and execute queries in several other languages, as illustrated in Fig.~\ref{fig:multilingual}.

\section{Conclusion}

We presented \coolname{} as an effective solution to the open-set multimodal 3D mapping problem.
The zero-shot nature of our method enables reasoning over a significantly broad range of concepts; leveraging off-the-shelf foundation features for open-set perception.
We evaluate our approach on in-house and established datasets, and on two real robotic systems (a manipulator and a self-driving vehicle).
Our results indicate several promising avenues for integrating foundation models trained over web-scale data with traditional mapping systems to enable zero-shot, open-set, and multimodal perception.

\textbf{Limitations}: The key limitations of our method are threefold. First, \coolname{} operates over dense maps, often comprising millions of 3D points over an apartment-scale scene, and augments each point with high-dimensional concept embeddings, requiring large amounts of memory.
Second, our pixel-aligned features are predominantly centered around foreground objects, lack compositionality, and do not understand concepts of negation.
Research efforts mitigating these shortcomings~\cite{li2022composing,lewis2022does} are ripe for exploration.
Third, we anticipate \coolname{} to inherit the limitations and biases of foundation models~\cite{fmodels, bender2021dangers}, warranting further investigations for potential harm as well as research into AI safety and alignment~\cite{gabriel2020artificial, yudkowsky2016ai}.

\section*{Appendix}

\setcounter{section}{0}
\setcounter{equation}{0}
\setcounter{figure}{0}
\setcounter{table}{0}

\renewcommand{\thesection}{A\arabic{section}}
\renewcommand{\thefigure}{A.\arabic{figure}}
\renewcommand{\thetable}{A.\arabic{table}}
\renewcommand{\theequation}{A.\arabic{equation}}

\section{Contribution statement}

\textbf{Krishna Murthy} conceived the idea and led the project. Was responsible for much of the code development and wrote sections of the paper. Curated and annotated the UnCoCo dataset and helped with the tabletop robot experiments.

\textbf{Ali Kuwajerwala} organized the initial brainstorming session that kick-started this project. Collected parts of the real-world test data, curated image and text queries, implemented various features necessary for conducting experiments, created several graphics including the explainer video, and wrote sections of the paper.

\textbf{Qiao Gu} implemented key components of the system, including 3D fusion modules and 2D-3D semantic segmentation evaluation protocols. Ran important ablation experiments and contributed to the writing of the paper.

\textbf{Mohd Omama} conducted all of the autonomous driving experiments and contributed significantly to producing figures and videos for the paper.

\textbf{Tao Chen} led the tabletop rearrangement experiments and played an instrumental role in generating insights that led to the creation of the UnCoCo dataset.

\textbf{Alaa Maalouf} made it possible to run the \coolname{} system in real-time, by quantizing and tracing all foundation models used in this work.

\textbf{Shuang Li} led experiments integrating large language models as planners and contributed to the writing of the paper.

\textbf{Ganesh Iyer} made significant research contributions to the gradslam framework (prior to joining Amazon) and follow-up work. Helped write sections of the paper.

\textbf{Soroush Saryazdi} was a key contributor to the gradSLAM library that \coolname{} was built upon (work done prior to joining Matician).

\textbf{Nikhil Keetha} contributed to several negative results that helped shape the direction of the research and wrote sections of the paper.

\textbf{Ayush Tewari} and \textbf{Celso de Melo} participated in multiple brainstorming sessions that helped shape \coolname{}.

\textbf{Josh Tenenbaum} provided valuable cognitive science perspectives and constructive skepticism, which informed the direction of the research and drew our attention towards potential issues (and interesting follow-up directions).

\textbf{Madhava Krishna} advised on the real-world autonomous driving experiments and suggested a crucial restructuring of the paper.

\textbf{Liam Paull, Florian Shkurti}, and \textbf{Antonio Torralba} were involved in brainstorming and critical review throughout the project, always asking the hard questions that led to key research insights. Wrote and proofread sections of the paper.

\section{Acknowledgements}

This project was supported in part (KM, AT, JBT) by the Army Research Laboratory (grant W911NF1820218), and from the office of naval research (ONR) via the multidisciplinary university research initiatives (MURI) program. KM and JBT also acknowledge funding support from a university engagement research grant from Lockheed Martin Corporation. The authors thank Tamar Rott Shaham, Chuang Gan, Joanna Materzynska, Songyou Peng, and Toni Rosinol for discussions and useful feedback over the course of this project. FS and LP thank NSERC for funding support. LP acknowledges support from the Canada CIFAR AI Chairs program. We thank Pulkit Agrawal for providing us with a UR5e robot for real-world experiments.

\section{3D feature fusion details}

For \textbf{indoor datasets} (ScanNet~\cite{scannet}, Replica~\cite{replica}, AI2-THOR~\cite{ai2thor}, ICL~\cite{icldataset}, UnCoCo), we implement our 3D feature fusion algorithm on top of the $\nabla$SLAM dense reconstruction framework.
By doing so, we leverage the point-based fusion technique proposed in ~\cite{pointfusionbase}, ensuring that points on nearby surface patches share the same \textit{surfel}, decreasing the overall number of map elements, and also increasing the effective number of pixels that contribute to each map element.
Another benefit we obtain is the ease of integration with PyTorch~\cite{pytorch}, which interfaces with a large number of foundation models.
For pointfusion, we use the default hyperparameters as suggested in ~\cite{pointfusionbase}, i.e., a distance threshold of 5 cm (on positions) and an angular threshold of 20 degrees (on normals) is used to discard noisy correspondences.

On \textbf{outdoor datasets} (SemanticKITTI~\cite{semantickitti}, self-captured autonomous driving sequences), we incrementally register pointclouds into a global frame using the LegoLOAM~\cite{shan2018lego} technique for odometry estimation.
We first compute all image points that have a valid map point by projecting the lidar depths onto the image plane.
We associate the features at these pixels with the corresponding 3D locations.

\section{Pixel-aligned feature extraction}

We use instance segmentation models from Mask2Former~\cite{cheng2021mask2former}; specifically the Swin-L backbone pretrained for image classification on ImageNet and subsequently finetuned for class-agnostic instance proposal generation on the COCO dataset.
Note that we only use the class-agnostic instance proposal generator; and do not use any of the subsequent modules, which are explicitly trained with instance segmentation ground-truth.
This results in 100 mask proposals per image.
We allow each each pixel to recieve fused features from multiple overlapping or redundant masks.
This is achieved by a running normalization whenever features from a new mask are assigned to a pixel.

\section{Foundation models used}

We use two broad classes of foundation models: DINO (and associated vision transformers)~\cite{dino}, and CLIP (and variants)~\cite{clip}.

\textbf{Vision transformer variants} include various DINO backbones implemented in ~\cite{dino}, as well as several vision-transformer variants explored in ~\cite{dinofts}.

\textbf{CLIP models used}: We use open-source CLIP models from the OpenAI CLIP~\cite{openai-clip-github} and OpenCLIP~\cite{open-clip-github} packages. We also use the publicly available AudioCLIP~\cite{audioclip}, trained on AudioSet~\cite{audioset}.

\begin{table*}[!ht]
    \centering
    \begin{tabular}{ll}
    \toprule
    \textbf{Sequence ID} & \textbf{Set of objects in the sequence} \\
    \midrule
    Seq 03 & Steel pouring mug, Ceramic coffee mug, Plastic banana, Windex spray bottle, SoftScrub \\
    Seq 04 & Baymax plush toy, Green caterpillar plush toy, Hedgehog plush toy \\
    Seq 05 & Hand-drill, Wooden spatula, Large lego block, Whiteboard marker \\
    Seq 06 & Plastic apple, Plastic grapes, Bottle of Vitamin E pills, Orange-colored bowl, Purple toy \\
    Seq 07 & Whisk, Spatula, Prongs, Silicone pastry brush \\
    Seq 08 & Paper cup, Spindrift can, Can of evaporated milk, Goldfish cheddar snack \\
    Seq 09 & Orange plastic cup, Paper cup, Steed pouring cup, Block of wood \\
    Seq 10 & Game of Bandu, Reacher grabber, Kitchen towel roll, Lysol wipes \\
    Seq 11 & Garbage bags, Cheetos, Steel measuring cup, Face mask \\
    Seq 12 & Coffee beans, Energy bar, Salted peanuts, Paper plates, Sugar sachet \\
    Seq 13 & Red hat, Magic candle, Molecule toy, Alligator toy, Blue frisbee \\
    Seq 14 & GoPro, Measuring tape, Scissors, Smartphone \\
    Seq 15 & Post-it notes, Black ceramic mug, Mustard, Tomato Ketchup \\
    Seq 16 & Bowl filled with sea shells, Ceramic vase, Large stapler \\
    Seq 17 & Stuffed mouse toy, Playing cube, Algorithms textbook, USB stick \\
    Seq 18 & USB adapter, NVIDIA Jetson board, Battery, Steel ruler \\
    Seq 19 & Compact Disk, Hard drive box, Teddy Bear, Inflatable brain toy \\
    Seq 20 & 3D glasses, Spray bottle, Charger block, Purell bottle \\
    \bottomrule
    \end{tabular}
    \caption{List of objects from the UnCoCo sequences used for evaluation. The first two sequences (not listed here) were used for tuning hyperparameters.}
    \label{suppl:table:uncoco-object-list}
    \vspace{-0.4cm}
\end{table*}

\section{More details on our experiment setup}

In all evaluations presented in the paper, we focus only on foreground objects, ignoring five background classes (\textit{wall, floor, ceiling, door, window} for indoor scenes, and \textit{road, sidewalk, building}) for outdoor scenes.

\textbf{ScanNet}: We note that most sequences from the ScanNet dataset suffer from motion blur artifacts, devoiding several interesting objects of texture; or are small rooms devoid of interesting objects.
We inspected every sequence (and each frame therein) over the ScanNet validation set, and identified the following sequences as being at least the scale of a one-bedroom apartment, and not suffering motion blur: \texttt{scene0011}, \texttt{scene0050}, \texttt{scene0231}, \texttt{scene0378}, \texttt{scene0518}.
We also use \texttt{scene0084} and \texttt{scene0168} for debugging and tuning our reconstruction system (and consequently, these two scenes are left out of our evaluation set).

\textbf{Replica}: We evaluate on the following 8 replica scenes \texttt{office0}, \texttt{office1}, \texttt{office2}, \texttt{office3}, \texttt{office4}, \texttt{room0}, \texttt{room1}, \texttt{room2}.

\textbf{Other datasets}: We also qualitatively evaluate our mapping system over all sequences from the ICL~\cite{icldataset} and on floorplans 9 and 402 from the AI2-THOR~\cite{ai2thor} simulator.
On SemanticKITTI~\cite{semantickitti}, we evaluate on all image frames containing at least one foreground object.

\section{Details of the UnCoCo dataset}

\begin{figure*}[!t]
    \centering
    \includegraphics[width=\linewidth]{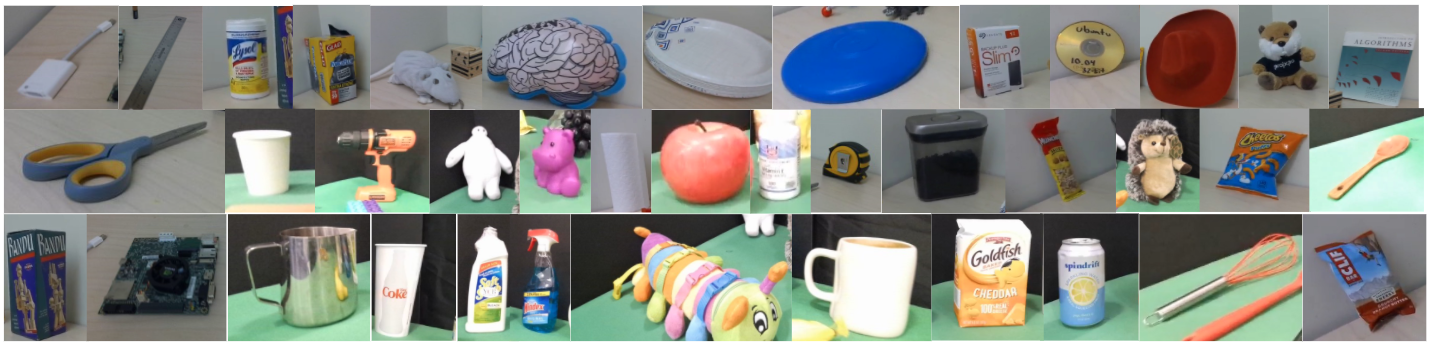}
    \caption{A subset of objects from the UnCoCo dataset. The dataset includes commonly found objects in homes and workplaces captured in a tabletop setting. Each object is annotated with 2D and 3D segmentation masks, and multimodal queries.}
    \label{suppl:fig:uncoco-objects}
    \vspace{-0.4cm}
\end{figure*}

The UnCoCo dataset comprises 78 commonly found objects in homes and workplaces, captured on tabletop settings over 20 RGB-D sequences. A subset of objects from UnCoCo is visualized in Fig.~\ref{suppl:fig:uncoco-objects}.
Of the captured 20 sequences, one was used for tuning parameters of the RGB-D reconstruction algorithm~\cite{gradslam} and another was used for tuning hyperparameters (thresholds over cosine similarity scores); so we exclude these two sequences from evaluation.

We list the set of objects available across the 18 validation sequences in Table~\ref{suppl:table:uncoco-object-list}.

\section{Details on 3D spatial query modules}

\definecolor{codeblue}{rgb}{0.38039216, 0.61568627, .8       }
\definecolor{codepink}{rgb}{1.        , 0.41568627, 0.83529412}
\definecolor{codegray}{rgb}{0.5,0.5,0.5}
\definecolor{codepurple}{rgb}{0.58,0,0.82}
\definecolor{backcolour}{rgb}{0.98,0.98,0.98}

\lstdefinestyle{mystyle}{
    backgroundcolor=\color{backcolour},   
    commentstyle=\color{codeblue},
    keywordstyle=\color{codepink},
    numberstyle=\tiny\color{codegray},
    stringstyle=\color{codepurple},
    basicstyle=\ttfamily\scriptsize,
    breakatwhitespace=false,         
    breaklines=true,                 
    captionpos=b,                    
    keepspaces=true,                 
    numbers=left,                    
    numbersep=5pt,                  
    showspaces=false,                
    showstringspaces=false,
    showtabs=false,                  
    tabsize=2
}
\lstset{style=mystyle}

\lstdefinelanguage{gptprompt}{
  basicstyle=\ttfamily\footnotesize,
  keywords = {howFar, isToTheRight, isToTheLeft, onTopOf, under, isBigger, canFitInside, isContained},
  keywordstyle=\color{red},
  keywords = [2]{Query, Response},
  keywordstyle = [2]\color{blue},
}

In Sec.~\ref{sec:spatial-query-building-blocks}, we described our approach to handling 3D spatial reasoning queries.
As investigated in~\cite{subramanian2022reclip,liu2021learning,li2022composing}, CLIP does not inherently capture spatial relationships or compositions.
We therefore implement a set of primitive 3D spatial comparator (3DSC) modules, and rely on a large language model~\cite{gpt3} (LLM) for parsing a natural language query into the function signatures (function name and input arguments) of the corresponding 3DSC.
In particular, we use the \texttt{text-davinci-003} model from OpenAI, with the default settings (temperature of 0.7, maximum length of 256, and TopP equalling 1).
We first condition the LLM by presenting a list of available 3DSCs and a brief natural language description of their behavior, followed by one example query and response.
Here is the exact text prompt used.
\begin{lstlisting}[language=gptprompt, caption=Base prompt used to condition for spatial queries]
Here is a set of available functions:
1. howFar(object1, object2): returns the distance between object1 and object2
2. isToTheRight(object1, object2): returns true if object1 is to the right of object2
3. isToTheLeft(object1, object2): returns true if object1 is to the left of object2
4. isContained(object1, object2): returns true if object1 is contained in object2
5. onTopOf(object1, object2): returns true if object1 is on top of object 2
6. under(object1, object2): returns true if object1 is underneath object 2
7. isBigger(object1, object2): returns true if object1 is bigger than object2
8. canFitInside(object1, object2): returns true if object1 can fit inside object2
Parse the provided queries into one of the above function formats.

Query: How close is the chair from the sofa?
Response: howFar(chair, sofa)
\end{lstlisting}

Once conditioned with this prompt, the model is able to parse language queries into function signatures. We directly execute these function signatures as is. We find LLMs to be very effective at parsing: of the 100 queries we used, each one was parsed correctly.
Shown below are a few outputs.
\begin{lstlisting}[language=gptprompt, caption=Sample outputs from the LLM after conditioning]
Query: Is the bread inside the bowl?
Response: isContained(bread, bowl)

Query: Is the apple on the table?
Response: onTopOf(apple, table)

Query: How far is the sanitizer from the door?
Response: howFar(sanitizer, door)

Query: I want to know the distance between the door and the window.
Response: howFar(door, window)

Query: Where is the closest restroom from my location?
Response: howFar(restroom, my location)

Query: I want to grab a can of soda and put this into a bag.
Response: canFitInside(soda, bag)

Query: Is the soda inside the bag?
Response: isContained(soda, bag)
\end{lstlisting}

\bibliographystyle{unsrt}
\bibliography{references}

\end{document}